\documentclass[letterpaper]{article} 
\usepackage{aaai24}  
\usepackage{times}  
\usepackage{helvet}  
\usepackage{courier}  
\usepackage[hyphens]{url}  
\usepackage{graphicx} 
\usepackage{natbib}  
\usepackage{caption} 
\usepackage{algorithm}
\usepackage{algorithmic}

\usepackage{amssymb}
\usepackage{bm}

\usepackage{newfloat}
\usepackage{listings}
\DeclareCaptionStyle{ruled}{labelfont=normalfont,labelsep=colon,strut=off} 
\floatstyle{ruled}
\newfloat{listing}{tb}{lst}{}
\floatname{listing}{Listing}
\usepackage{xcolor}
\title{High-Fidelity Diffusion-based Image Editing}
\author{
    Chen Hou\textsuperscript{\rm 1},
    Guoqiang Wei\textsuperscript{\rm 2},
    Zhibo Chen\textsuperscript{\rm 1}\thanks{Corresponding author.}
}
\affiliations{
    \textsuperscript{\rm 1}University of Science and Technology of China,  
    \textsuperscript{\rm 2}ByteDance Research\\

    houchen@mail.ustc.edu.cn, weiguoqiang.9@bytedance.com, chenzhibo@ustc.edu.cn
}

\usepackage{bibentry}

\begin{document}

\maketitle

\begin{abstract}
Diffusion models have attained remarkable success in the domains of image generation and editing. It is widely recognized that employing larger inversion and denoising steps in diffusion model leads to improved image reconstruction quality. However, the editing performance of diffusion models tends to be no more satisfactory even with increasing denoising steps.
The deficiency in editing could be attributed to the conditional Markovian property of the editing process, where errors accumulate throughout denoising steps. 
To tackle this challenge, we first propose an innovative framework where a rectifier module is incorporated to modulate diffusion model weights with residual features, thereby providing compensatory information to bridge the fidelity gap. 
Furthermore, we introduce a novel learning paradigm aimed at minimizing error propagation during the editing process, which trains the editing procedure in a manner similar to denoising score-matching. 
Extensive experiments demonstrate that our proposed framework and training strategy achieve high-fidelity reconstruction and editing results across various levels of denoising steps, meanwhile exhibits exceptional performance in terms of both quantitative metric and qualitative assessments. 
Moreover, we explore our model's generalization through several applications like image-to-image translation and out-of-domain image editing.
\end{abstract}

\section{Introduction}

\begin{figure}[t]
    \centering
    \includegraphics[width=0.9\columnwidth]{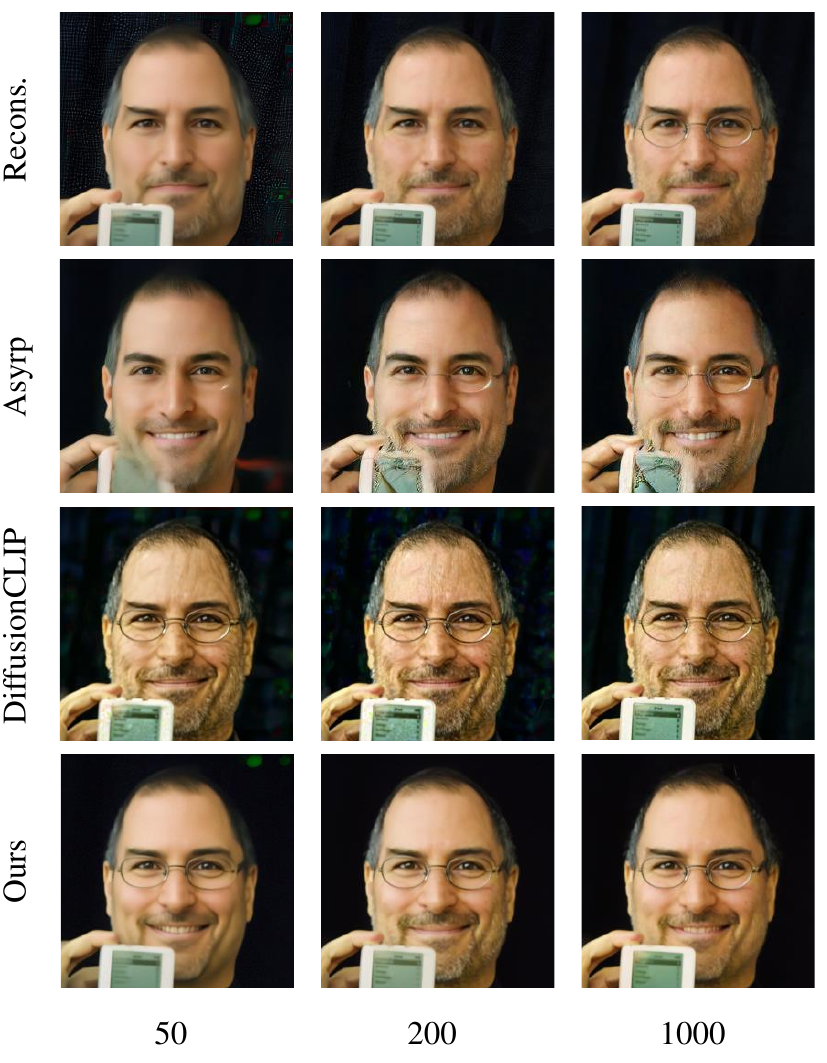}
    \caption{Reconstruction and editing results under various levels of inversion and denoising steps. While increasing steps makes reconstruction nearly perfect, the outcomes of editing still remain far from satisfactory (attribute: smiling).}
    \label{fig1}
\end{figure}

As a rising star of generative models, tremendous works of diffusion models \cite{ho2020denoising, song2020score} have been exploded in recent years. Except for works focusing on optimizing diffusion algorithm itself \cite{nichol2021improved, song2020denoising}, others devote to studying how to add controllable conditions to diffusion models, including adding image guidance \cite{choi2021ilvr}, classifier guidance \cite{dhariwal2021diffusion, avrahami2022blended}, using representation learning \cite{kwon2022diffusion} or additional networks \cite{rombach2022high, zhang2023adding}. These methods then inspire series of applications based on diffusion models, like image inpainting \cite{lugmayr2022repaint}, image translation \cite{meng2021sdedit}, super resolution \cite{ho2022cascaded} and image editing \cite{nichol2021glide, kwon2022diffusion, hertz2022prompt}. 

Existing work on image editing based on diffusion models could be roughly divided into two categories. One is through image guidance \cite{nichol2021glide, yang2023paint}, these methods take advantage of diffusion model's image-level noisy maps, and achieve editing by adding pixel-wise control through denoising process. But the disadvantages are that these methods need either mask \cite{avrahami2022blended}, estimating mask \cite{couairon2022diffedit} or segmentation map \cite{matsunaga2022fine} to get fine control of images, besides, they are not suitable for semantic editing and the editing directions are usually heterogeneous. Other methods manipulate images via internal representation of diffusion, by exploring semantic latent space \cite{kwon2022diffusion, preechakul2022diffusion}, or finetuning model parameters \cite{kim2022diffusionclip, kawar2023imagic}. These method don't need mask as constraints, and except some of them only handle single image and corresponding text prompt as input \cite{hertz2022prompt, kawar2023imagic}, they get editing directions in good properties that are homogenoues, linear and robust \cite{kwon2022diffusion}. Despite the superiorities, these methods, due to they offset the denoising process following editing directions, often cause changes in irrelevant attributes, and the details of image will also be lost or distorted. It should be noted that for reconstructions, increasing diffusion steps could do nearly perfect reconstruction for most images, but this doesn't hold true in the case of editing. The deficiency in editing could be attributed to its conditional Markovian property, leading to error accumulating and amplifing \cite{mokady2023null}. Fig.1 shows some reconstruction and editing results with various levels of diffuse and denoise steps from 50 (the lowest common steps used to save time) to 1000 (the highest steps adopted to train original DDPM), the editing attribute is smiling. As illustrated, reconstruction attains nearly perfect results with increasing steps, whereas for editing, the outcomes are still far from satisfactory.

To solve these problems, we first analyze why diffusion models suffer from distorted reconstructions or edits, and how these problems could be alleviated. Following that, we propose a designed framework and develop an effective training strategy to resolve these issues. Firstly we do this by adding a rectifier into diffusion model to fill the fidelity gap during denoising process. The rectifier is a hypernetwork \cite{david2016hypernetworks} that encodes the residual feature of original image and each step's estimation, at every step, it learns to predict the offsets of convolutional filters' weights for diffusion model's corresponding layers, providing compensated information for high-fidelity reconstruction. Secondly, to further reduce the propagation of error during editing process, we introduce a new paradigm for training editing based on diffusion models. Unlike previous methods who adopt Markov-like training strategies that make error accumulation \cite{kim2022diffusionclip, kwon2022diffusion}, we train editing in a way like denoising score matching \cite{song2020score} which is wildly used in training diffusion models \cite{ho2020denoising, song2020score, lipman2022flow}. This restrains the trajectory deviation caused by editing not to accumulate, and effectively improves the faithfulness of edited results. Extensive experiments show that our method produces high-fidelity reconstruction and editing results without retraining diffusion model itself, especially for out-of-domain images.

To summarize, the main contributions are:
\begin{itemize}
\item We propose an innovative framework to achieve high-fidelity reconstruction and editing based on pretrained diffusion model, where a rectifier is incorporated to modulate model weights with residual features, providing compensated information for bridging the fidelity gap.

\item To further reduce error propagation during editing, we propose a new learning paradigm where editing is trained in a manner similar to denoising score-matching. This prevents denoising trajectory from accumulated deviation, effectively improves the fidelity of edited results.

\end{itemize}

\section{Related Work}

\begin{figure*}[t]
    \centering
    \includegraphics[width=0.9\textwidth]{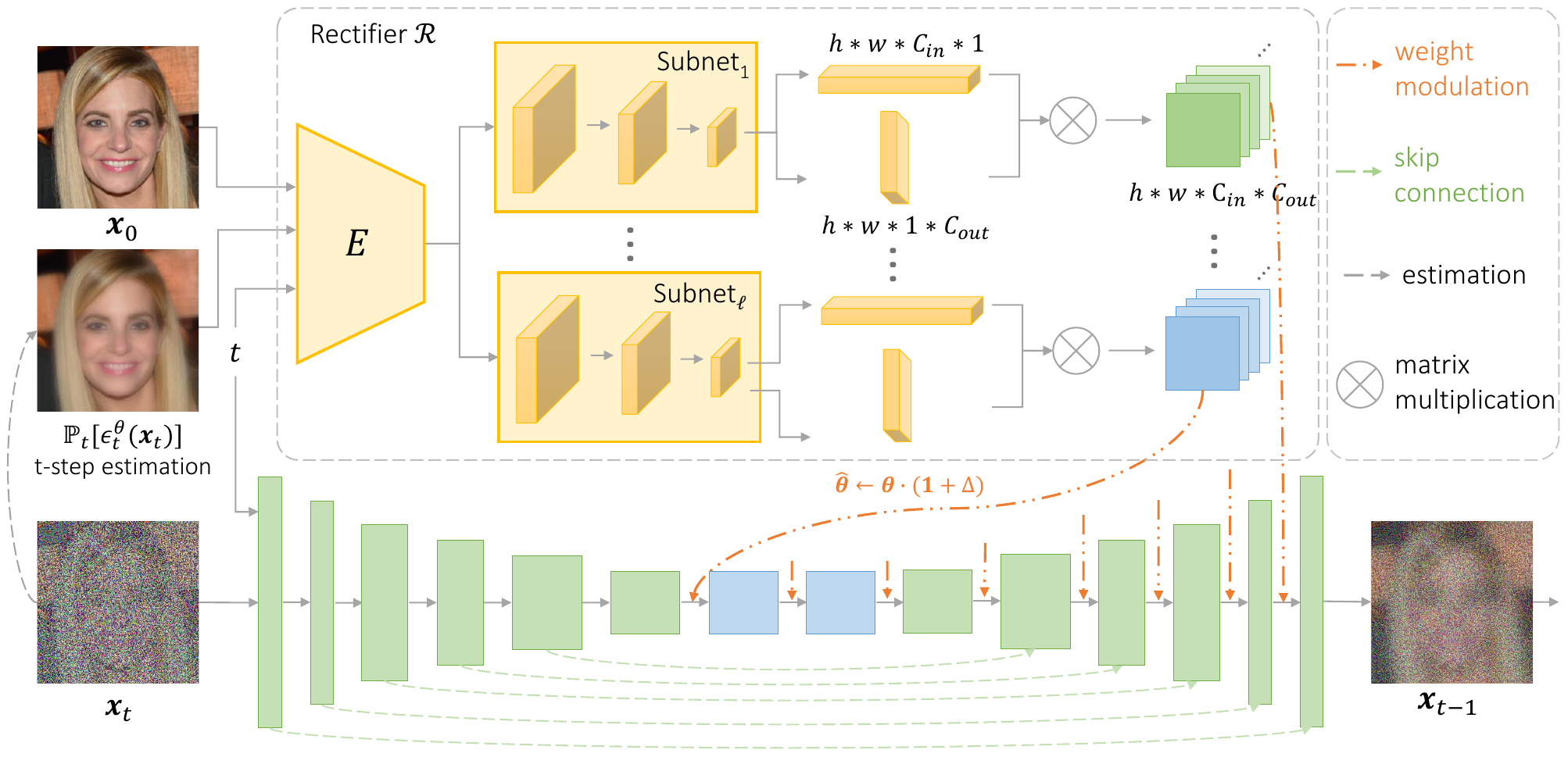}
    \caption{Overview of our proposed rectifier framework. The rectifier is a hypernetwork consisting of a global encoder and multiple subnet branches. 
    It takes as input the original image $\bm{x}_0$ and the estimation at each step ($\mathbb{P}_t[\bm{\epsilon}_t^\theta(\bm{x}_t)]$), targets to modulate the degraded residual features into offset weights, providing compensated information for high-fidelity reconstruction. 
    We select the middle and up-sampling blocks of U-Net for modulate, considering that these blocks contain both high-level semantic information and low-level details. 
    We also employ separable convolution to reduce the amount of generated parameters.}
    \label{fig2}
\end{figure*}

\subsection{Image Editing with Diffusion Models}
The most intuitive way of using diffusion model for editing is to utilize the intermediate noisy maps generated during denoising process. These maps have the same resolutions as output images, making it convenient to directly add pixel-wise controls for manipulation, and their noisy property retains randomness for generation diversities. Many works take this advantage and apply it in various tasks like semantic editing \cite{choi2021ilvr}, image translation \cite{meng2021sdedit}, inpainting \cite{lugmayr2022repaint}, and pixel-level editing with mask \cite{nichol2021glide,yang2023paint,avrahami2022blended}. Some other methods explore the influence of internal representation to attribute editing, instead of changing sampling process, they change the diffusion model itself by exploring the semantic latent inside \cite{kwon2022diffusion}, or finetuning the model to adapt editing tasks \cite{kim2022diffusionclip, hertz2022prompt, kawar2023imagic}. These methods could get homogenous and robust editing directions without the help of mask, but often suffer from distortion and low-fidelity. Besides interfering the denoising process or finetuning diffusion model, some methods take a novel yet different path to achieve editing by modulating the initial noise \cite{mao2023guided}. There are also some novel methods who offer customized text control by inverting images into textual tokens \cite{gal2022image, mokady2023null}.

\subsection{High-Fidelity Inversion of GANs}
Unlike the natural inversion capability exists in diffusion models \cite{song2020denoising}, GANs \cite{goodfellow2020generative} need to do inversion by encoder \cite{richardson2021encoding}, optimization \cite{abdal2020image2stylegan++} or a combination of both \cite{zhu2020domain}. Poor fidelity of inversion and reconstruction leads to the distortion-editability trade-off in GAN-based editing tasks \cite{tov2021designing}. That is, good editing directions often lead to bad distortions and vice versa, it's hard to keep distortion and editing results both satisfying. Many works resolve this problem by improving inversion fidelity of GANs. Restyle \cite{alaluf2021restyle} achieves this goal through iterative refine the residual of latent code. StyleRes \cite{pehlivan2023styleres} transforms the residual of feature maps instead of images into editing branch, and propose a cycle-consistency loss to retain input details. HFGI \cite{wang2022high} instead adaptively aligns the distortion map then fuses it into generator's internal feature maps, similar idea is also presented in ReGANIE \cite{li2023reganie}. While most works often keep generator weight unchanged, there are other methods like HyperStyle \cite{alaluf2022hyperstyle} who chooses to finetune generator parameters. Motivated by these methods, while considering the particularity of diffusion model relative to GAN, we propose a high-fidelity framework adapted for diffusion models. 

\section{Methodology}

In this section, we start by explaining why diffusion models suffer from distorted reconstructions and edits. Then we will elaborate on how these issues could be alleviated, followed by the introduction of our method.

\subsection{High-Fidelity Problem in Diffusion Models}

Reconstructions of diffusion model are not always perfect. As claimed in PDAE \cite{zhang2022unsupervised}, 
the major reason of these imperfections 
is there exists a clear gap between the predicted posterior mean and the true one. Compared to reconstruction, editing deviates denoising trajectory thus leads to more error accumulation \cite{mokady2023null}, while \cite{ho2022classifier} also figures out that the effect induced by editing condition (such as classifier or conditioned text) will be amplified during denoising process, making editing a harder task than merely reconstruction.

According to \cite{zhang2022unsupervised}, some prior knowledge about $x_0$ introduced to the reverse process will help reduce the gap and achieve better reconstruction. From this perspective, classifier-guidance \cite{dhariwal2021diffusion} method can be seen as utilizing the class information to fill this gap, via shifting the predicted posterior mean with an extra item computed by the classifier's gradient \cite{zhang2022unsupervised}. PDAE also proves that this is equivalent to shifting the noise predicted by the model, thus they use an additional network predicting noise shift to make up for the information loss.

\subsection{Hypernetwork as Rectifier}
While these methods point out how to compensate for the information gap, their reconstructions are still far from high-fidelity, and resorting to external network or classifier also makes it difficult to generalize to semantic editing tasks which  mainly relies on diffusion's internal representations. In this work, we propose a framework where a rectifier is incorporated to modulate residual features into offset weights, providing compensated information to help pretrained diffusion model achieving high-fidelity reconstructions. The framework is illustrated in Fig. \ref{fig2}. 
Our rectifier is a hypernetwork \cite{david2016hypernetworks} which takes as input every step's estimation and original image, expected to exploit the degradated residual features to fill the fidelity gap.
The inputs first pass through a global encoder, then transformed by a series of sub-nets to generate layer-wise modulation. 
We choose to modulate the middle and up-sampling blocks of U-Net \cite{ronneberger2015u}, considering they contain both high-level semantic information and low-level details. Furthermore, without the interference of other representations like classifier, our framework is highly adaptive for semantic editing tasks, and is easier to generalize to other diffusion-based downstream tasks.

For parameter modulation, we generate offsets for all convolutional layers' kernel weights, instead of regenerating them from scratch. This can preserve prior knowledge of pretrained diffusion model as much as possible \cite{alaluf2022hyperstyle}. Specifically, at time step $t$, the rectifier $\bm{\mathcal{R}}$ takes in the original image $\bm{x}_{0}$ and the prediction result using $\bm{x}_{t}$, then outputs weight offsets $\Delta_{t}$ for $\ell$-th layer of U-Net which are then assigned to each channel $i$ of $j$-th filter:
\begin{figure}[t]
    \centering
    \includegraphics[width=1\columnwidth]{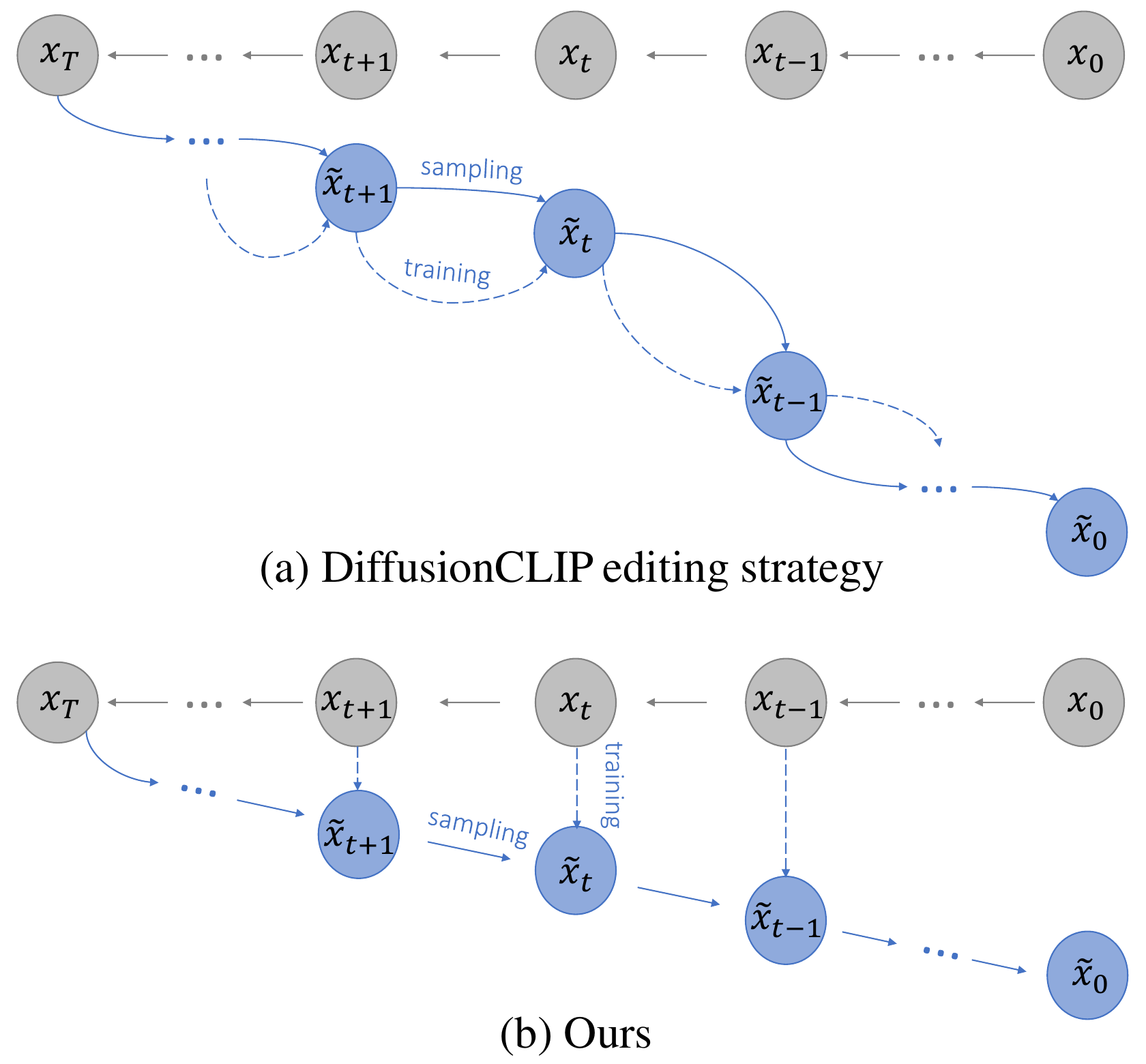}
    \caption{Editing training strategy. Instead of shifting from previous edited results in a Markovian style used in DiffusionCLIP (a), which may lead to error propagation, we start from the original trajectory at each step to find editing direction (b), further alleviating error accumulation caused in editing process.}
    \label{fig3}
\end{figure}

\begin{equation}
    \Delta_{\ell,t}^{i,j} := \bm{\mathcal{R}}(\bm{x}_{0}, \mathbb{P}_t[\bm{\epsilon}_t^\theta(\bm{x}_t)], t),
\end{equation}
where $\bm{\epsilon}_t^\theta$ represents the noise estimation at time step $t$ with parameter $\theta$. $\mathbb{P}_t[\bm{\epsilon}_t^\theta(\bm{x}_t)]=(\bm{x}_t-\sqrt{1-\alpha_t}\bm{\epsilon}_t^\theta)/\sqrt{\alpha_t}$ refers to the estimation of $\bm{x}_0$ using $\bm{x}_t$ defined in DDIM \cite{song2020denoising}, and $\alpha_t$ denotes the transformed variance noise schedule used in DDPM \cite{ho2020denoising}. 
The kernel weight is modulated as:
\begin{equation}
    \hat{\theta}_{\ell,t}^{i,j} := \theta_{\ell,t}^{i,j} \cdot (1+\Delta_{\ell,t}^{i,j}).
\end{equation}

Considering the huge cost of estimating weight offsets for all selected layers,
we employ separable convolution \cite{alaluf2022hyperstyle} to cut down the amount of parameters generated. Rather than predicting offsets for every filter of every channel (which requires $\sum_{\ell}h * w * C_{in} * C_{out}$ parameters generated in total), we decompose it into two parts: $h*w*C_{in}*1$ and $h*w*1*C_{out}$, their product is taken as the final output. In this way, the number of parameters is reduced to $\sum_{\ell}(h*w*C_{in}*1 + h*w*1*C_{out})$.
This significantly reduces memory usage of the network, while not affecting its capability too much. For loss function, we choose noise fitting loss as our training objective:

\begin{equation}
    \mathcal{L}_{rec} := \mathbb{E}_{t,\bm{x}_0,\bm{\epsilon}}\left [ \left \| \bm{\epsilon} - \bm{\epsilon}_{t}^{\hat{\theta}}(\bm{x}_t) \right \|_{2}^{2}\right]. 
\end{equation}

It is rational to consider other loss functions, like $\ell_1 ~loss$, which is commonly used in GAN finetuning tasks \cite{alaluf2022hyperstyle, wang2022high}. We also validate the effectiveness of these loss functions for
diffusion models, and the relevant results are shown in supplementary material.

\subsection{Training Editing like Score Matching}

As claimed before, compared with reconstruction, editing is more challenging and more susceptible to causing distortion during denoising process, owing to the error accumulation introduced by input condition. How to diminish these impact and keep high-fidelity for editing is a critical issue we should consider next. Current methods train editing in a Markovian way \cite{kim2022diffusionclip, kwon2022diffusion}, in which case the deviation of denoising trajectory will gradually accumulate, 
leading to 
irrelevant attributes change, 
details loss or distortion (Fig.1). To alleviate this problem and further reduce the error propagation in editing process, we propose a training strategy that trains editing in a manner similar to denoising score matching \cite{song2020score}. Our editing training strategy is depicted in Fig.3. The inspiration is drawn from the training strategy of
diffusion model \cite{ho2020denoising} and score-based generative model \cite{song2020score}.
Instead of drifting from previous edited results in a Markovian way, 
we instead take 
the original trajectory as the starting point to find editing directions for each step. 
This eschews the accumulation of the 
the deviations caused by editing,
and further reduces the error propagated in editing process. Specifically, we reuse the rectifier to modulate model's weights served for editing, which can also be interpreted as shifting the output distribution of the entire model along the direction of attribute change.

\begin{algorithm}[t]
    \caption{Editing Training Strategy}
    \label{alg:algorithm}
    \begin{algorithmic}[1]
        \REPEAT 
            \STATE $\bm{x}_0 \sim q(\bm{x}_0)$\\
            \STATE $t \sim \mathrm{Uniform}(\{1,...,T\})$\\
            \STATE $\bm{\epsilon} \sim \mathcal{N}(\bm{0},\bm{I})$\\
            \STATE $\bm{x}_t = \sqrt{\bar{\alpha}_t}\bm{x}_0 + \sqrt{1-\bar{\alpha}_t}\bm{\epsilon}$\\
            \STATE $\tilde{\theta} \gets \theta \cdot (1+\bm{\mathcal{R}}(\bm{x}_{0}, \mathbb{P}_t[\bm{\epsilon}_t^\theta(\bm{x}_t)], t))$\\
            \STATE Take gradient descent step on\\
                \qquad $\nabla_{\mathcal{R}} \mathcal{L}_{direction}(\mathbb{P}_t[\bm{\epsilon}_t^{\tilde{\theta}}(\bm{x}_t)], t_{tar};\bm{x}_{0},t_{src})$\\
                \qquad $\nabla_{\mathcal{R}} \mathcal{L}_{\ell_1}(\mathbb{P}_t[\bm{\epsilon}_t^{\tilde{\theta}}(\bm{x}_t)], \bm{x}_0)$
        \UNTIL{converged}
    \end{algorithmic}
\end{algorithm}

Another advantage of our editing training strategy is that we do not need to specify any heuristic-defined parameters to fit different attributes. It should be remarked here that for methods like Asyrp \cite{kwon2022diffusion} and DiffusionCLIP \cite{kim2022diffusionclip}, neither of them employs editing through the entire denoising process. Asyrp halts editing prematurely and adds stochastic noise then to boost preceived quality, while DiffusionCLIP does not inverse images into complete noise for preserving their morphologies. Setting these parameters meticulously for every separate attribute is intricate and bothersome. Our method though, starts from pure noise and traverses the process throughly to get editing results, no extra settings are needed. 

We incorporate the directional CLIP loss \cite{gal2022stylegan} to train the editing process. Specifically, given the source image $\bm{x}_{src}$ and text $t_{src}$ as well as the target image $\bm{x}_{tar}$ and text $t_{tar}$, we can calculate the feature directions encoded by CLIP's image encodeer $E_I$ and text encoder $E_T$, \textit{i.e.}, $\Delta I = E_I(\bm{x}_{tar}) - E_I(\bm{x}_{src})$ and $\Delta T=E_T(t_{tar})-E_T(t_{src})$. The directional CLIP loss aims to align the image change $\Delta I$ and text change $\Delta T$, which could be formulated as: 

\begin{equation}
    \mathcal{L}_{direction}(\bm{x}_{tar}, t_{tar};\bm{x}_{src},t_{src}) := 1 - \frac{\left \langle \Delta I,\Delta T \right \rangle }{\left \| \Delta I \right \| \left \| \Delta T \right \| }.
\end{equation}

Motivated by \cite{kim2022diffusionclip}, we also introduce another $\ell_1$ loss as a regularizer to circumvent the change in irrelevant attributes:
\begin{equation}
    \mathcal{L}_{\ell_1}(\bm{x}_{tar}, \bm{x}_{src}) := \left \| \bm{x}_{tar} - \bm{x}_{src} \right \|.
\end{equation}

Our final loss function for training editing is:

\begin{equation}
    \mathcal{L}_{edit} := \lambda_{CLIP}\mathcal{L}_{direction} + \lambda_{recon} \mathcal{L}_{\ell_1}.
\end{equation}

Training of editing is established upon the foundation of model pretrained by the rectifier part. During inference, we still use the same sampling procedure as DDPM \cite{ho2020denoising}, but with the modulated model that leads to corresponding attribute change. Our training strategy is elucidated in Algorithm 1.

\section{Experiments}

\subsection{Implementation Details}
We conduct experiments on FFHQ \cite{karras2019style}, CelebA-HQ \cite{karras2017progressive}, AFHQ-dog \cite{choi2020stargan}, METFACES \cite{karras2020training}, LSUN-church/-bedroom \cite{yu2015lsun} datasets with the outcomes of various levels of steps, and all pretrained models are kept frozen. Note that due to the separable convolution used in rectifier, our model is GPU-efficient and are able to complete all training tasks on a single RTX 3090TI GPU.

\subsection{Reconstructions}
We present both quantitative and qualitative evaluations of image reconstruction. 
We conduct our rectifier on several backbones with various datasets, and the quantitative results are shown in Table \ref{tab:1}. 
iDDPM \cite{nichol2021improved} is employed for human faces, and the metrics are calculated on 10,000 random sampled images from CelebA-HQ using model trained on FFHQ under 50 inversion and sampling steps. In terms of natural scenes, we use DDPM++ \cite{song2020score} as foundation model and implement on LSUN-Church \cite{yu2015lsun} with 20 steps.
It is worth noting that even though we do not train on these indicators and only train with noise fitting loss as Eq.(1), our method still outperforms original model under some of the reconstruction quality assessment criterias. We also test the average posterior mean gap $\left \| \Delta e \right \| ^2$, and it turns out our method reaches lower gap than original model. These results manifest our rectifier could bring quality improvement for model's overall output distribution, and indeed provides compensated information thus fills the fidelity gap.

Some qualitative samples are shown in Fig.4. With the help of rectifier, our reconstructions become robust to occlusions, illuminations, viewpoints, and performs better at both restoring coarse shapes and  preserving details. More visual results can be found in the supplementary materials.

\subsection{Editings}


\begin{table}[t]
    \centering
    \caption{Quantitative results of image reconstruction.}
    \label{tab:1}
    \renewcommand\arraystretch{1.5}
    \resizebox{1.0\linewidth}{!}{
    \begin{tabular}{l | c c c c c}
        \hline
        Method & $L_1 (\downarrow)$ & $L_2(\downarrow)$ & LPIPS $(\downarrow)$ & SSIM $(\uparrow)$ & $\left \| \Delta e \right \| ^2 (\downarrow)$\\
        \hline
        iDDPM & 0.090 & 0.016 & 0.150 & 0.95 & 6.6711e-3\\
        Ours & \textbf{0.085} & \textbf{0.014} & \textbf{0.150} & 0.94 & \textbf{6.6710e-3}\\
        \hline
        DDPM++ & 0.255 & 0.109 & 0.643 & 0.48 & 1.559e-2\\
        Ours & \textbf{0.254} & \textbf{0.108} & \textbf{0.642} & \textbf{0.48} & \textbf{1.558e-2}\\
         \hline
    \end{tabular}
    }
\end{table}

\begin{figure}[t]
    \centering
    \includegraphics[width=1\columnwidth]{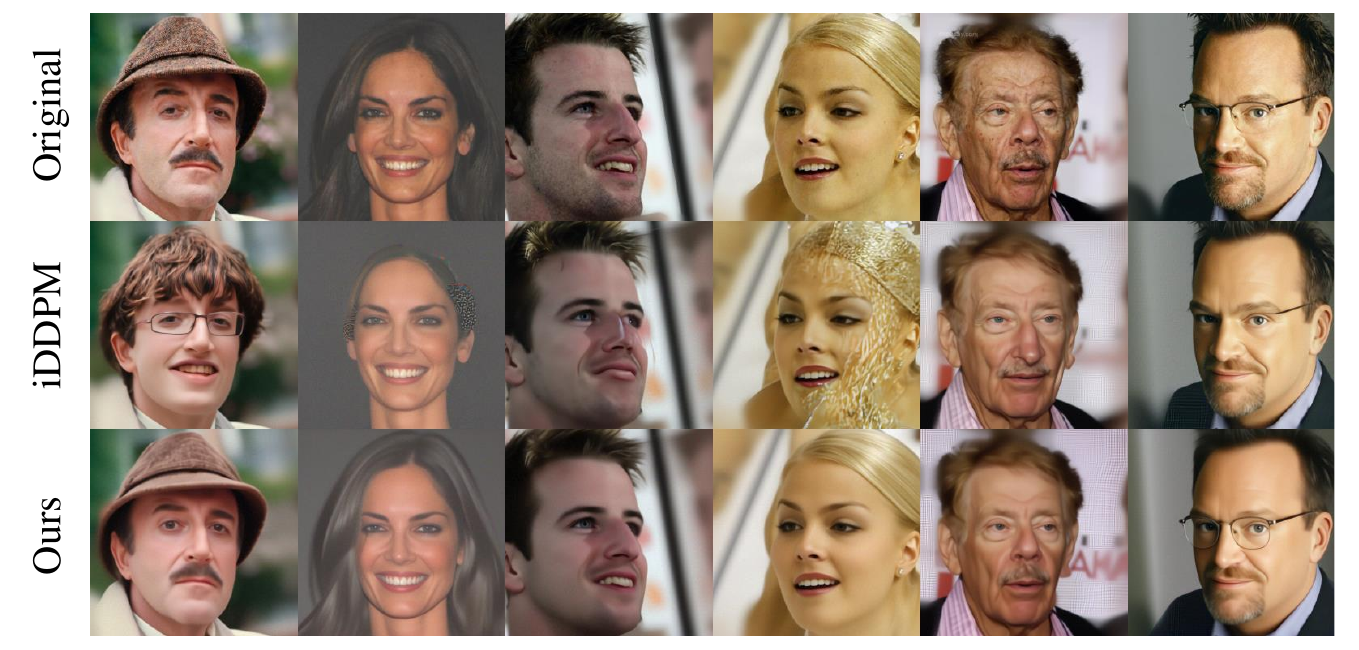}
    \caption{Comparison of reconstruction quality under 50 steps. Our method is more robust to occlusions (1st column), illuminations (2nd column), viewpoints (3rd and 4th columns), and performs better at restoring coarse shapes (5th column) and preserving fine details (6th column).}
    \label{fig4}
\end{figure}

\begin{figure*}[t]
    \centering
    \includegraphics[width=1\textwidth]{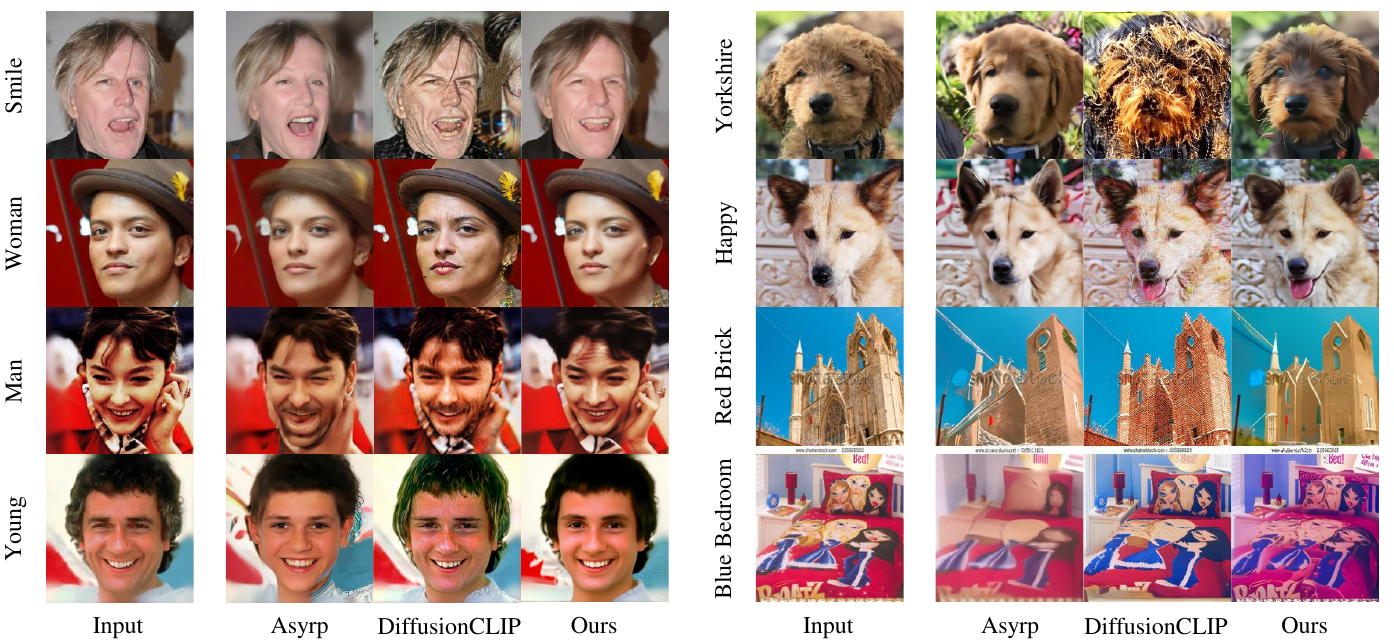}
    \caption{Editing qualitative comparisons. Our method delivers realistic edits while maintaining low distortion and high fidelity.}
    \label{fig5}
\end{figure*}

\begin{table}[t]
    \centering
    \caption{Quantitative comparisons of editing. We compare different methods with the identity similarity between original and edited images.}
    \label{tab:2}
    \renewcommand\arraystretch{1.1}
    \begin{tabular}{l | c c c }
        \hline
        Attribute & Asyrp & DiffusionCLIP & Ours \\
        \hline
        Man & 0.22$\pm$0.16 & 0.33$\pm$0.14 & \textbf{0.45$\pm$0.23} \\
        Pixar & 0.18$\pm$0.13 & 0.22$\pm$0.13 & \textbf{0.25$\pm$0.10}\\
         \hline
    \end{tabular}
\end{table}

For comparison of editing performance, we choose the representational editing methods based on diffusion backbones that retain state-of-the-art: Asyrp \cite{kwon2022diffusion}, DiffusionCLIP \cite{kim2022diffusionclip}. Among them, Asyrp leverages the deepest feature maps inside U-Net's bottleneck, treating it as diffusion model's semantic latent space to produce manipulations. DiffusionCLIP, on the other hand, directly finetunes the whole model for attaining editing results. We also conduct experiment on some image-guidance methods like GLIDE \cite{nichol2021glide} to test their abilities towards semantic editing, for which the results could be found in supplementary materials.

In a comparable manner to reconstruction part, both quantitative and qualitative evaluation of editing are exhibited here too. Fig.5 presents some qualitative comparisons towards different methods trained under 50 inversion and sampling steps. Like previously noted, neither of Asyrp nor DiffusionCLIP employs edit through the entire denoising process, Asyrp applies stochastic noise during final process to boost perceived quality, and DiffusionCLIP begins editing from intermediate noisy images for preserving their morphologies. Our method, though starts from complete noise and traverses the whole denoising process to edit, still carries out editing results with exceptional quality. Several illustrative examples are, for instance, elements like hat and ring are kept intact during semantic editing, along with the preservation of image's overall shape and background. Meanwhile, distortions or artifacts brought by conditional input text are avoided, and vital information loss is also alleviated.

Echoing what's previously mentioned, editing as a more challenging task compared to reconstruction, its quality does not tend to improve much even with increasing inversion and denoising steps, mainly due to the error accumulation introduced by input conditions. In order to reinforce this standpoint, we test the editing performance under various levels of inversion and sampling steps from 50 to 1000. The outcomes are highlighted in Fig.1. As can be observed, methods like Asyrp and DiffusionCLIP who use Markovian training strategy fail to produce realistic and faithful editing results, even with larger steps. Asyrp losts many essential information like the iPod in hand and the glasses. DiffusionCLIP benefits some details from its incomplete noise inversion, yet still leads to distortions and artifacts. Our method attains vivid editing results regardless the number of steps, meanwhile maintaining high-fidelity performance in preserving vital information and details.

We offer quantitative results as well. Given original images and their edited versions, we calculate the identity similarity using CurricularFace \cite{huang2020curricularface}, which grants us the capability to validate how identity are preserved before and after editing. Two attributes: man and pixar are evaluated, and all other methods are tested under official checkpoints. Table 2 showcases the results. It is evident from the table that out method obtains the highest identity similarity score among these attributes. Random sampled images used to calculate identity similarity and more details are shown in supplementary materials.

\subsection{Ablation Study}

\begin{figure}[t]
    \centering
    \includegraphics[width=1\columnwidth]{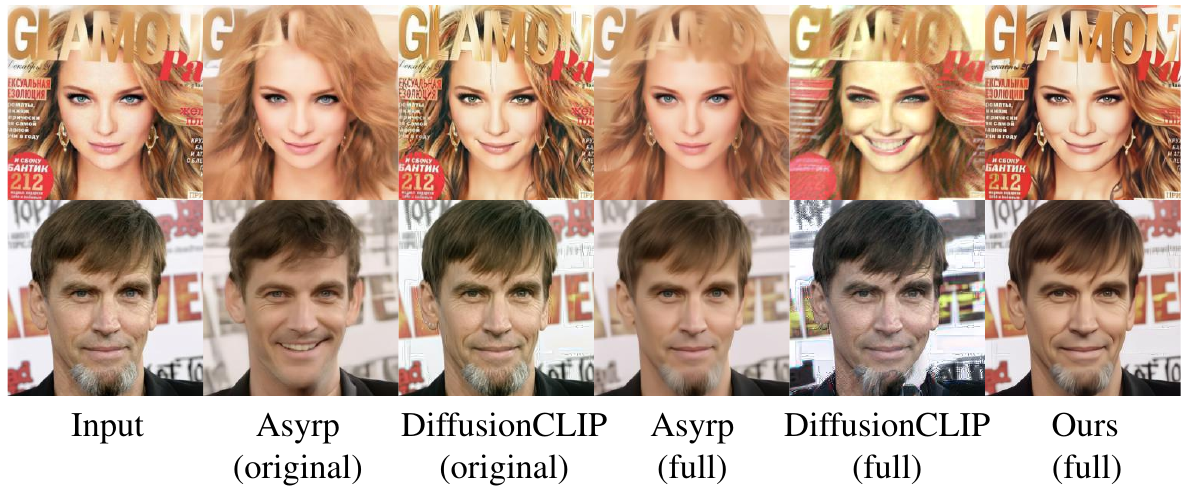}
    \caption{Effects of different editing training strategies. Different methods are evaluated across various ranges of editing intervals, "original" denotes default configuration, while "full" refers to editing through the entire denoising process. View with better clarity when zoomed-in.}
    \label{fig6}
\end{figure}

\subsubsection{Effect of Editing Training Strategy}

\begin{figure}[t]
    \centering
    \includegraphics[width=1\columnwidth]{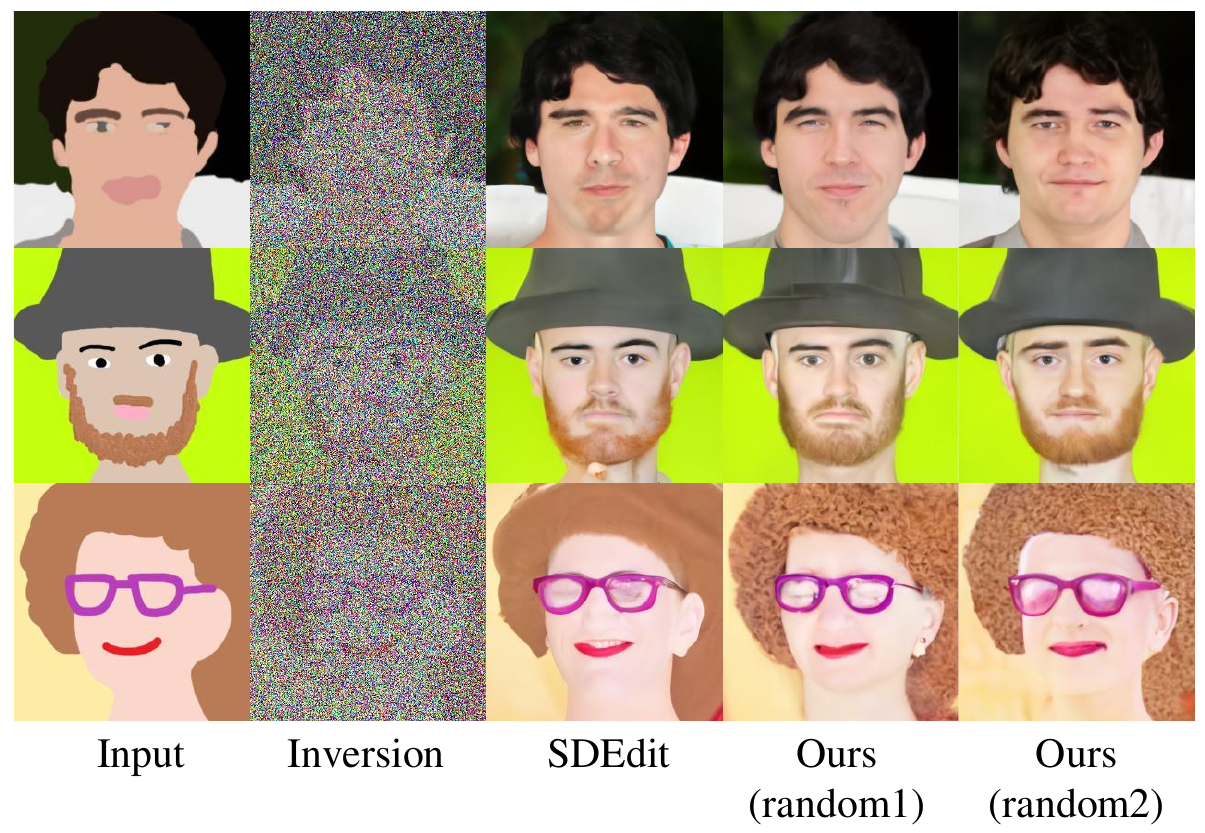}
    \caption{The influence of incorporating rectifier into SDEdit. The rectifier makes translation results  more lifelike and realistic, as well as exhibiting richer texture and details. No extra domain specific training are employed.}
    \label{fig7}
\end{figure}

\begin{figure}[t]
    \centering
    \includegraphics[width=1\columnwidth]{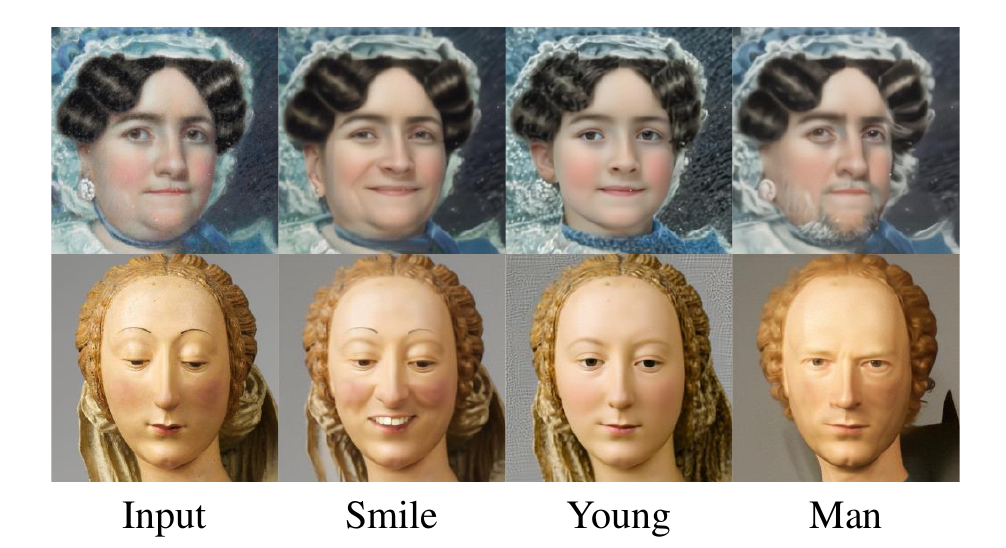}
    \caption{Generalize our method to out-of-domain images. Our model trained only on FFHQ successfully adapts to images from METFACES,  performing well on oil paintings and sculptures which possess intricate and unique textures unseen in FFHQ.}
    \label{fig8}
\end{figure}

Doubts may arise regarding to whether our gain in editing comes from the rectifier or the training strategy. With the benefit of rectifier already proven in preceding part, we now focus on ablation studies to validate the effectiveness of our editing training strategy alone. Attribute "smiling" is chosen for evaluating various algorithms.  Fig.6 presents the performance of these algorithms. As shown, even without any enhancement from rectifier, our strategy still produces results with less distortion and information loss compared to others. This demonstrates our strategy could be employed as an independent and generalized training approach for editing tasks based on diffusion models. Furthermore, recall that neither Asyrp nor DiffusionCLIP implements editing through the entire process, they either stops prematurely or starts from incomplete noise. We thus investigate how they perform when applied to full range editing, denoted as "full" in Fig.6. It turns out that longer editing interval does not yield satisfying editing results. Especially for method like DiffusionCLIP, extending interval instead leads to loss of details and many artifacts. This observation proves again that improving the performance of editing based on diffusion model necessitates more than simply increasing the editing interval steps.

\subsection{Further Applications}

\subsubsection{Image to Image Translation}
The rectifier module can be incorporated into any pretrained diffusion models to enhance their quality for overall output distribution, indicating its potential for generalizing to various downstream tasks that utilize diffusion model as basis. One of these tasks involves images translation. SDEdit \cite{meng2021sdedit} firstly exploits the advantage of diffusion's stochasticity and the prior knowledge hidden in pretrained model, making translation task simple to achieve. Here, we perform image translation in the same way SDEdit does, but with rectifier integrated, in order to evaluate its capability generalizing to other tasks. Noted that no additional domain-specific training are employed in this scenario, and we only adopt the rectifier pretrained on FFHQ dataset. The results are shown in Fig.7. Benefiting from the rectifier, the translation results become more realistic and exhibit richer in texture and details (like the hat and the hair). This inspiring finding demonstrates that our rectifier module indeed learns to produce compensated information, and possesses the capability of generalizing to other downstream tasks, bringing further quality enhancement for them.

\subsubsection{Generalization on Out-of-Domain Images}

For a more extensive evaluation of how our method generalizes, we further test its performance on images from other domains. Here we select images from METFACES and use our method pretrained on FFHQ dataset to edit. These out-of-domain images including oil paintings with complicated texture and details, as well as sculptures that possesses unique tactile qualities. We find out that even without any adjustment or finetuning for the new domain, our model could give expected outcomes achieving dual advantages in both editing performance and fidelity preservation. As depicted in Fig.8, while obtaining realistic and faithful edits, our method preserves greatly the intricate details such as texture of clothing, style of hair, together with images' whole structures. This signifies our method could handle diverse images from various similar domains, without explicitly finetuning on it, demonstrating its strong generalization ability.

\section{Conclusions}
In this work, we propose an innovative method to achieve high-fidelity image reconstruction and editing based on diffusion models. We employ a rectifier to encode residual feature into modulated weight, bringing compensated information for filling the fidelity gap. Furthermore, we introduce an effective editing learning paradigm which trains editing in a way like denoising score-matching, preventing error accumulation during editing process. By leveraging the rectifier and the training paradigm, our method produces high-fidelity reconstruction and editing results regardless of inversion and sampling steps. Comprehensive experiments validates the effectiveness of our method, and shows its strong generalization ability for editing out-of-domain images, or improving quality for various downstream tasks based on diffusion models.

\section{Acknowledgement}
This work was supported partly by Natural Science Foundation of China (NSFC) under Grant 62371434, 62021001.

\bibliography{aaai24}

\clearpage

\begin{huge}
Supplementary Materials
\end{huge}

\section{A. Model Architectures}

Our rectifier consists of a global encoder and a series of subnet branches to generate layer-wise modulations. ResNet34 \cite{he2016deep} backbone is chosen as the global encoder. Inspired by how GAN-based methods cope with residual features \cite{alaluf2022hyperstyle, pehlivan2023styleres, alaluf2021restyle}, we concatenate the original image and its estimation of every step $t$, constructing a 6-channel input to pass through global encoder. We also evaluate the performance of substracting the two inputs to obtain a 3-channel input, but the results are not good as 6-channel concatenation. The output is then processed by each subnet branch, which contains multiple convolutional layers to decrease the feature map size to 1 $\times$ 1, following with two convolutions to generate two slim matrices separately: $h \times w \times C_{in} \times 1$ and $h\times w\times 1\times C_{out}$ , the product of which is finally taken as the output modulated weight offset. Sinusoidal time embedding is integrated into each subnet branch using the same approach as U-Net, enabling the model with ability to perceive temporal representation. Detailed model architecture of rectifier and subnet are shown in Fig.1.

\section{B. Training Details}

During the training of editing, we set both $\lambda_{CLIP}$ and $\lambda_{recon}$ to 1. Adam optimizer is used for both reconstruction and editing training. We observe that the performance is substantially related to optimizer's hyperparameter and learning schedule. We set weight decay to 1e-5, learning rate to 1e-3 and decreased by 0.9 every 5000 steps for reconstruction training. For editing, setting weight decay to 0, learning rate to 1e-3 and decreased by 0.9 every 10 steps works fine for most attributes.
For those prompts that cause severe changes, we set learning schedule to lower decrease speed to avoid distortions. Only around 100 images are needed for training each attribute.

\begin{figure}[t]
    \centering
    \includegraphics[width=1\columnwidth]{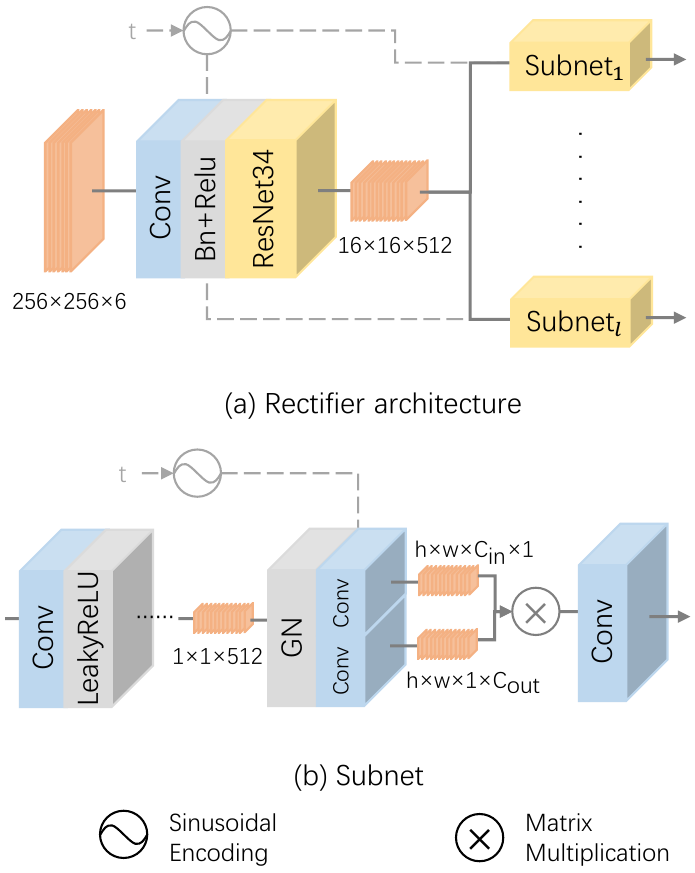}
    \caption{Detailed model architecture of rectifier (a) and its subnet (b).}
    \label{fig9}
\end{figure}

\section{C. Loss Function for Reconstruction Training}

Like most of fintuning methods based on diffusion models, we also employ the original noise fitting loss used in diffusion model \cite{ho2020denoising} itself for our reconstruction training. 
We also consider other loss functions like $\ell_1 ~loss$ which is commonly used in GAN's high-fidelity fintuning tasks. 
Here we validate the effectiveness of different loss functions, including noise fitting loss $e ~loss$, $\ell_1 ~loss$ and $\ell_1 ~loss$ with an output regularization $dw ~loss$ (output delta weights should be as small as possible).
Experiments are conducted under 50 inversion and sampling steps, the results of which are shown in Fig.2. From the figure we can see that $\ell_1 ~loss$ indeed restores the overall shape of input compared to original reconstruction, but the trade-off is that images become excessively smooth, along with loss of numerous details and texture, making the outcome not appeal like a real image (3rd row). The introduction of $dw ~loss$ could alleviate this problem, as it confines the changes on original diffusion model, yet it also leads to failure of some expected restoration (4th row). Despite the other two, $e ~loss$ achieves best results, for both recovering the overall shape and preserving the texture, 
it is surprisingly to see that $e ~loss$ accomplish so much alone.
Our exploration of loss function demonstrates that applying the loss function from previous methods (e.g. GAN-based methods) for finetuning diffusion models might not be the best choice, and the $e ~loss$ used for training diffusion models may contain more information than expected.

\begin{figure}[t]
    \centering
    \includegraphics[width=1\columnwidth]{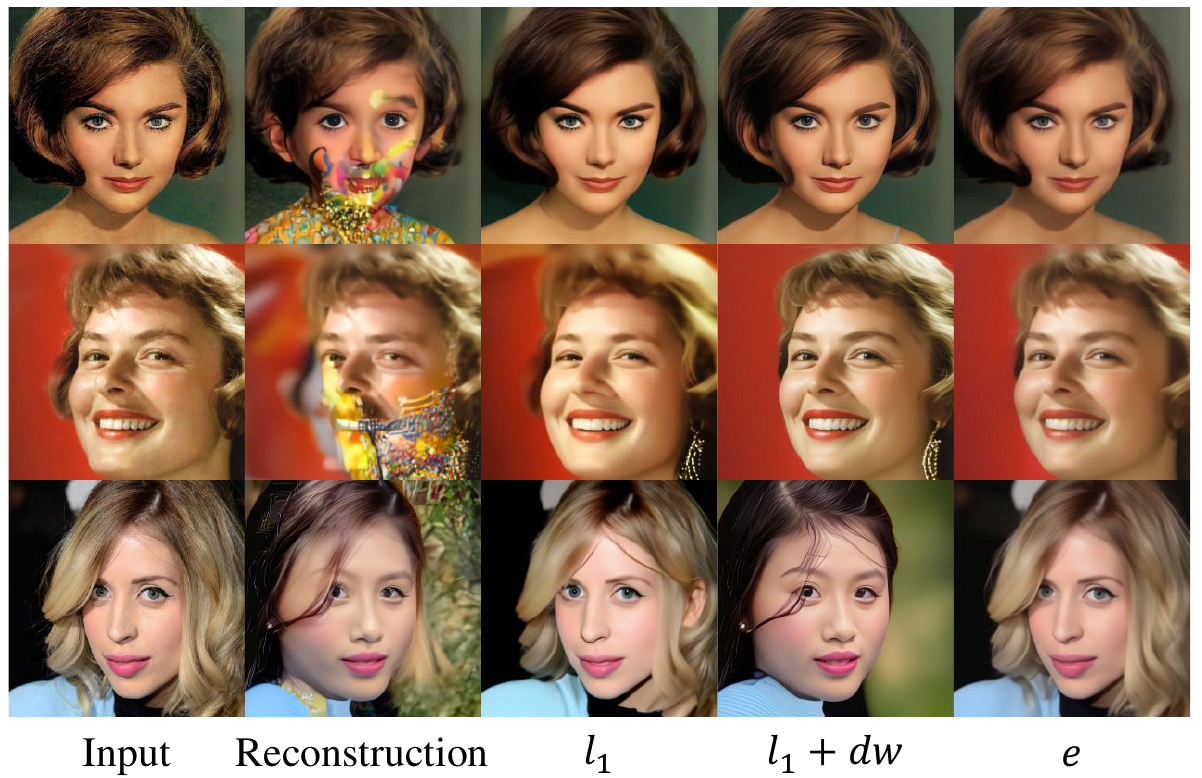}
    \caption{Reconstruction under different loss functions. $e ~loss$ accomplishes both restoring overall shapes and preserving details compared to other loss functions.}
    \label{fig10}
\end{figure}

\section{D. Hyper-parameters}

Explorations are done towards the influence of different ranges of $\lambda_{CLIP}$ and $\lambda_{recon}$ in Eq. (6), as illustrated in Fig.3. The results demonstrate that larger $\lambda_{recon}$ tends to retain original content, while larger $\lambda_{CLIP}$ inclines to alter the attributes. Therefore, for attributes requiring large changes, larger $\lambda_{CLIP}$ is recommended, and larger $\lambda_{recon}$ works better for those which needs to maintain identity.

\begin{figure}[t]
    \centering
    \includegraphics[width=1\columnwidth]{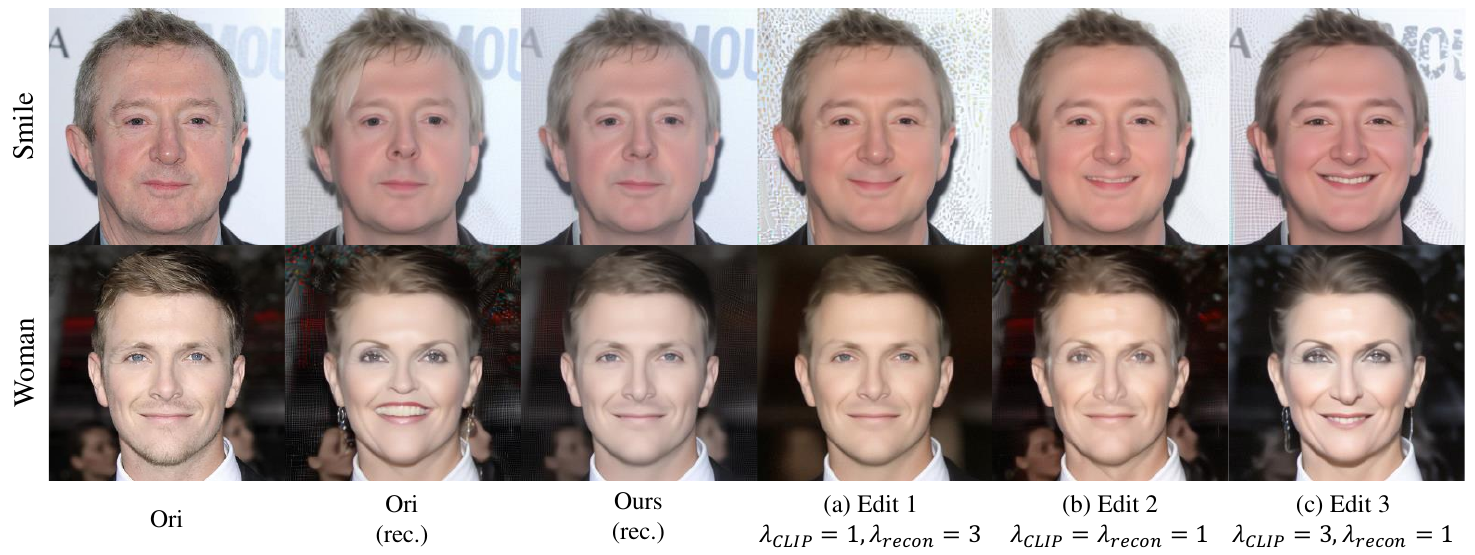}
    \caption{Exploration on hyper-parameters.}
    \label{fig11}
\end{figure}

\section{E. Editing Attributes Exploration}

During editing, we indeed notice some attributes  harder to train, resulting in subtle or irrelevant changes (\textit{e.g.}, \textit{angry}). We reckon that this might due to the low frequency for such attributes occurred in the text used for CLIP training. To confirm this, we calculate the occurrence frequency of some attributes in LAION-400M (available): 'smile': 198656, 'sad': 199474, 'angry': 21743. As a typical thorny attribute, \textit{angry} is notably less frequent, providing some support to our hypothesis.

\section{F. Additional Results}

1. Reconstructions and editings comparisons on CelebA-HQ dataset \cite{karras2017progressive} with representational based methods: Asyrp \cite{kwon2022diffusion} and DiffusionCLIP \cite{kim2022diffusionclip}. Evaluations are conducted under various inversion and denoising steps from 50 to 1000, Fig.4, Fig.5 and Fig.6 represent these results separately. Both DiffusionCLIP and our method are trained under same 100 images, while official Asyrp checkpoints are loaded for evaluation.

~\

2.  Reconstructions and editing comparisons on AFHQ-Dogs \cite{choi2020stargan} is shown in Fig.7. Besides two representational based editing methods mentioned above, we also test an image guidance based method GLIDE \cite{nichol2021glide} for AFHQ-Dogs dataset (pretrained GLIDE models on human faces are not released due to safety considerations), masks used in GLIDE inference are exhibited as well. 

~\

3. Image samples used in identity similarity calculation are displayed in Fig.8, we edit around 100 images with each method for evaluation.

~\

4. More results on further applications including image-to-image translation and out-of-domain image editing are shown in Fig.9 and Fig.10, respectively.

\begin{figure*}[t]
    \centering
    \includegraphics[width=0.9\textwidth]{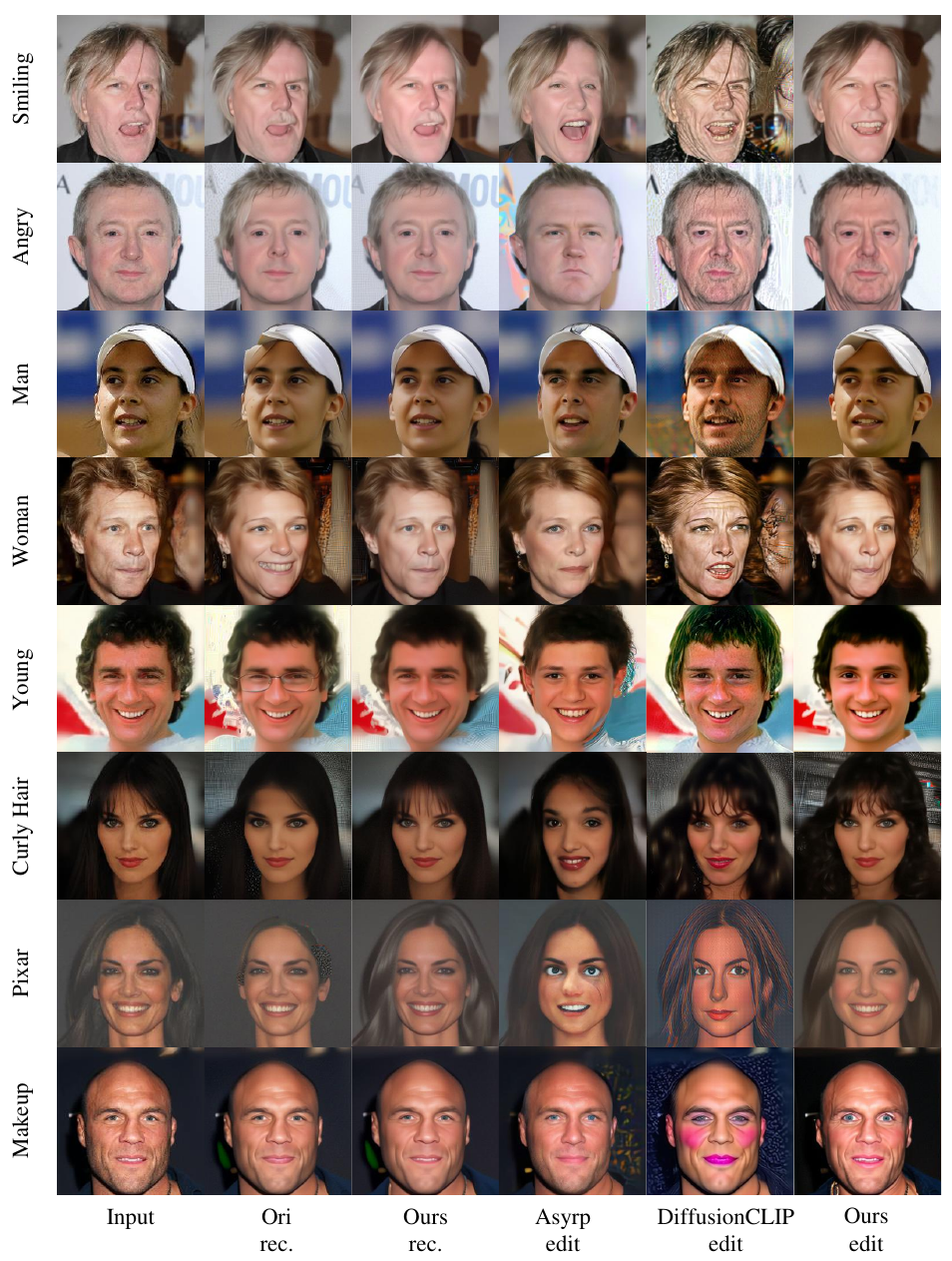}
    \caption{Visual comparisons of reconstructions and editings, under 50 inversion and denoising steps.}
    \label{fig12}
\end{figure*}

\begin{figure*}[t]
    \centering
    \includegraphics[width=0.8\textwidth]{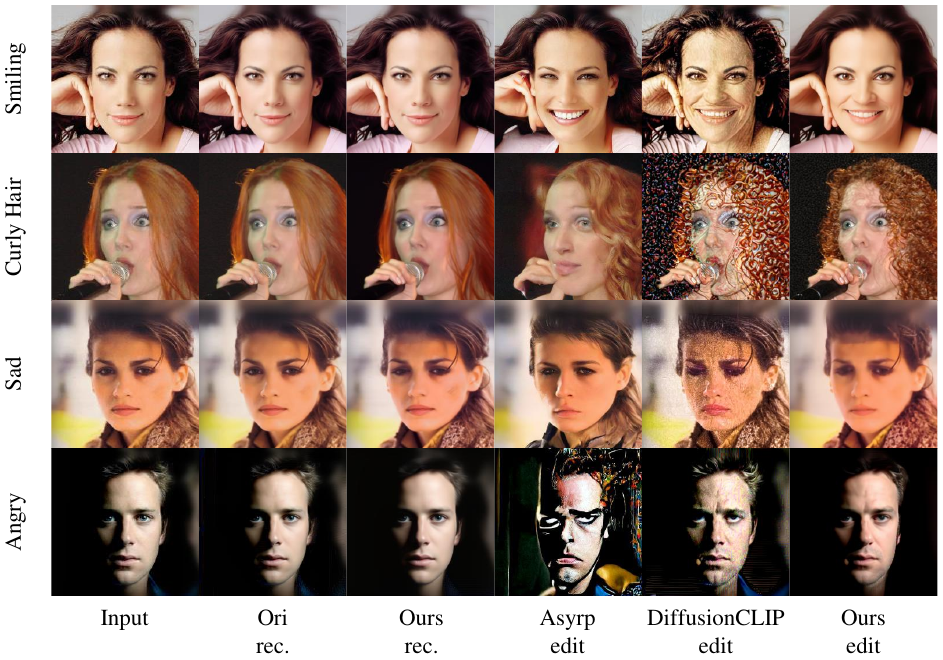}
    \caption{Visual comparisons of reconstructions and editings, under 200 inversion and denoising steps. As the reconstruction quality for most images are quite well over 200 steps, this could also be seen as an ablation study toward the effectiveness of our editing training strategy alone.}
    \label{fig13}
\end{figure*}

\begin{figure*}[t]
    \centering
    \includegraphics[width=0.8\textwidth]{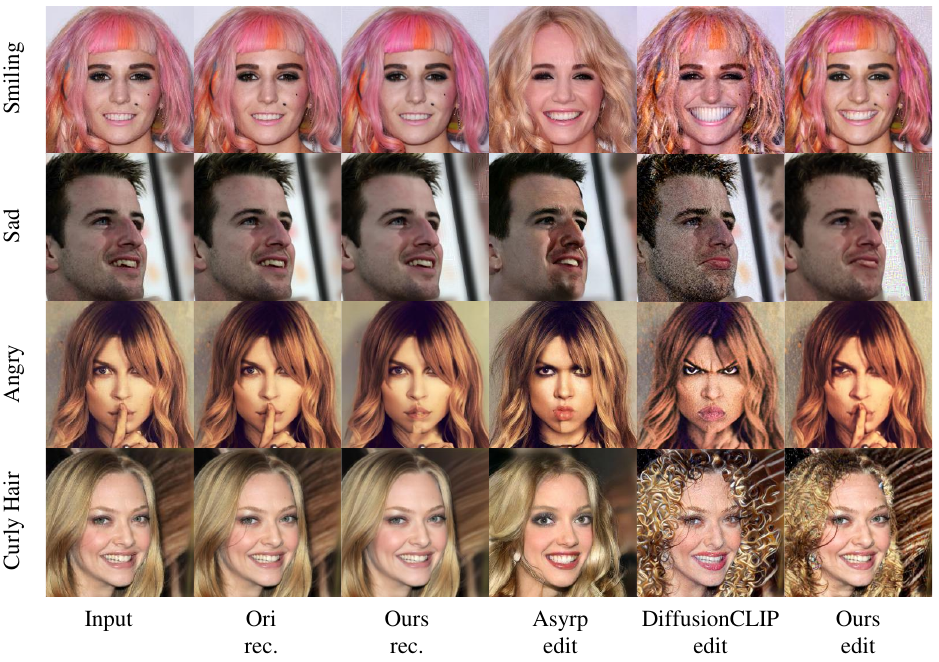}
    \caption{Visual comparisons of reconstructions and editings under 1000 steps.}
    \label{fig14}
\end{figure*}

\begin{figure*}[t]
    \centering
    \includegraphics[width=0.9\textwidth]{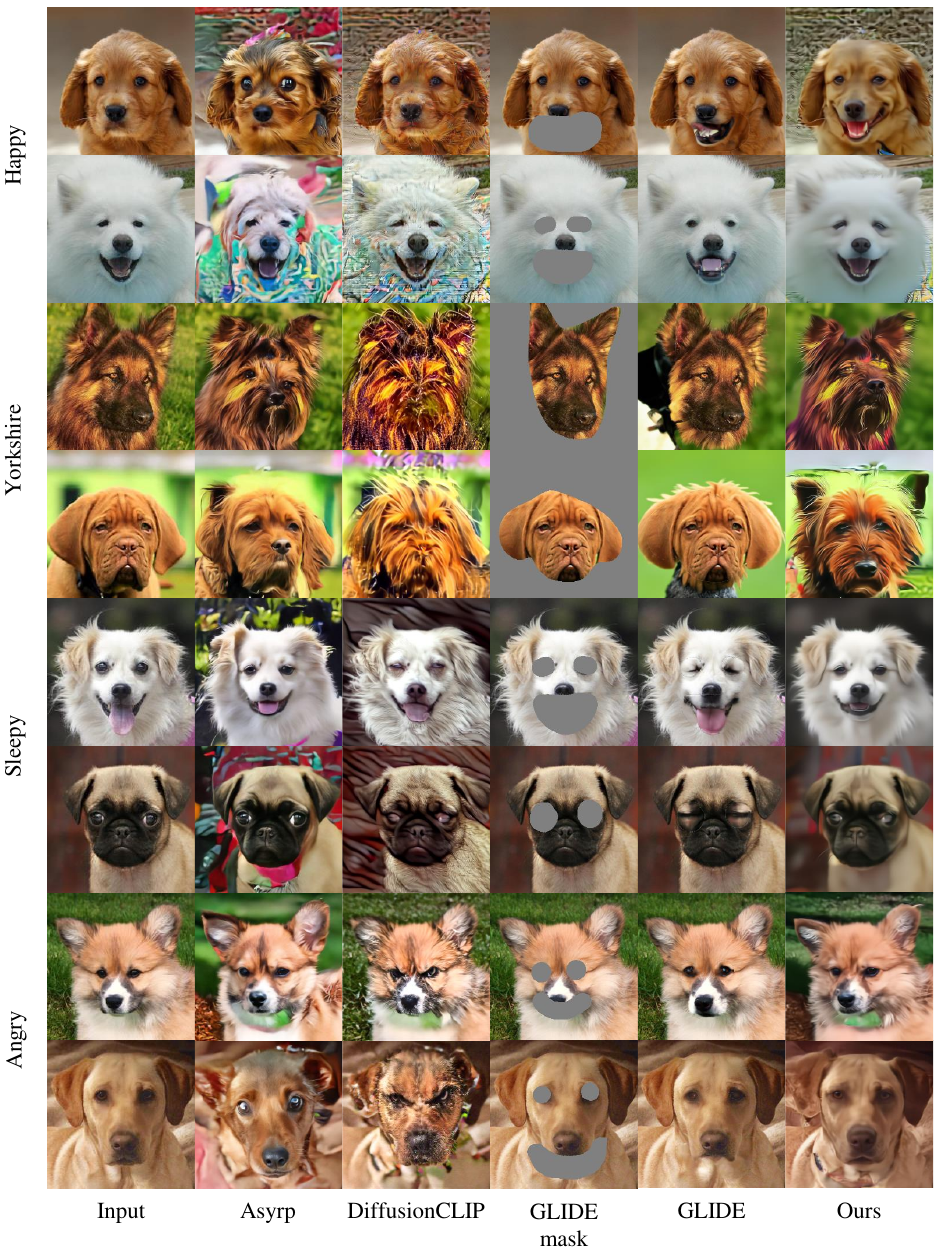}
    \caption{Visual comparisons of editings of AFHQ-Dogs. GLIDE achieves great detail preservation due to its mask mechanism, but sometimes causes inconsistency and unsatisfying results in semantic editing.}
    \label{fig15}
\end{figure*}

\begin{figure*}[t]
    \centering
    \includegraphics[width=1\textwidth]{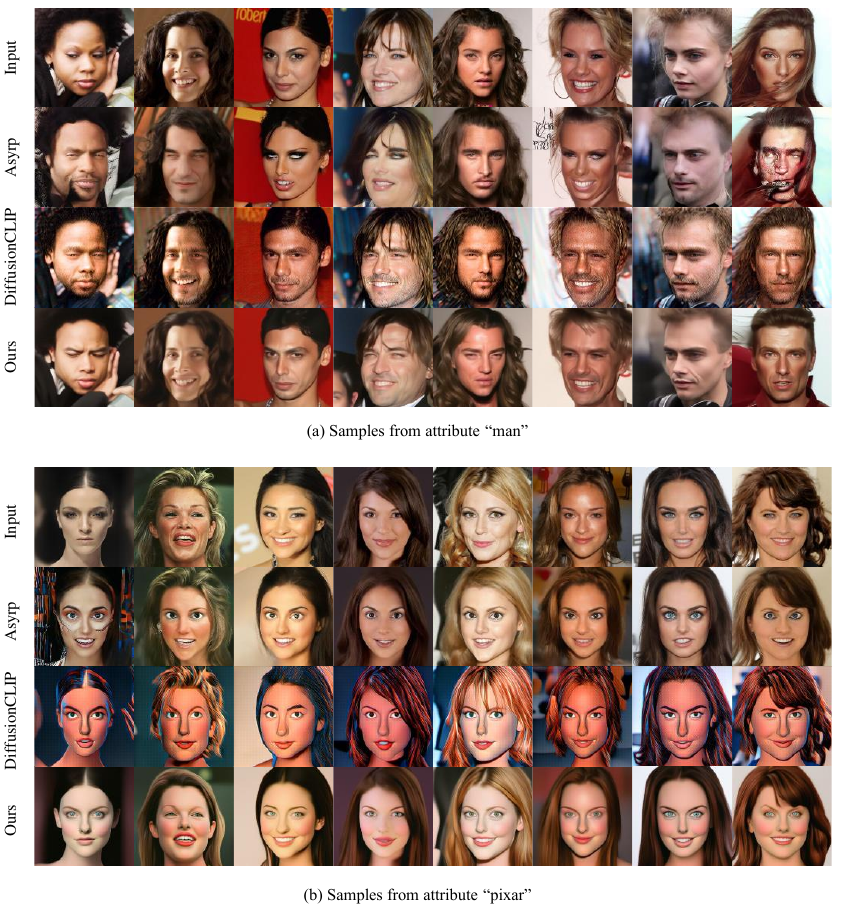}
    \caption{Image samples for calculating identity similarity.}
    \label{fig16}
\end{figure*}

\begin{figure*}[t]
    \centering
    \includegraphics[width=1\textwidth]{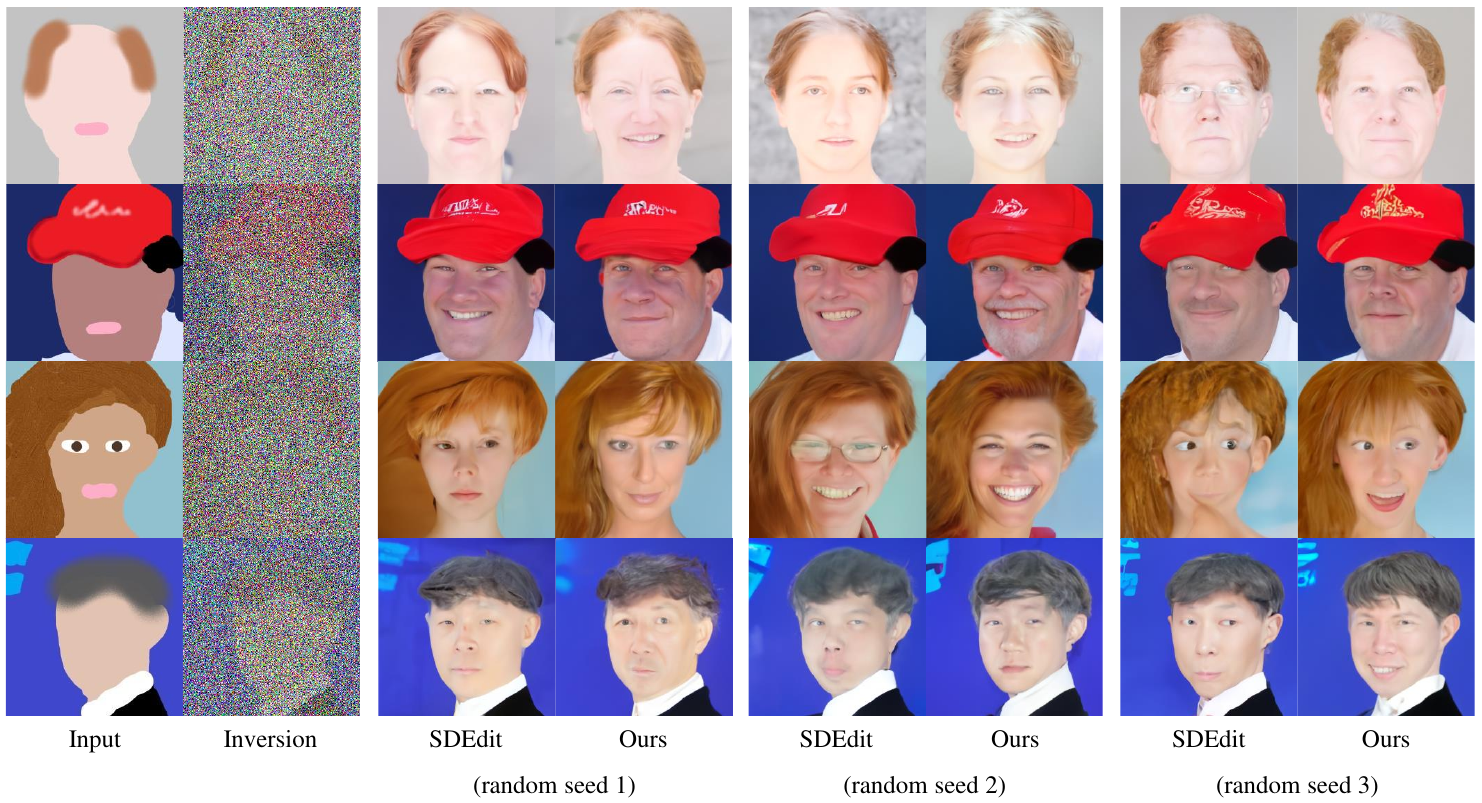}
    \caption{Applications on image-to-image translation, compared to SDEdit\cite{meng2021sdedit}. Every two columns of translation results share common random seed. }
    \label{fig17}
\end{figure*}

\begin{figure*}[t]
    \centering
    \includegraphics[width=1\textwidth]{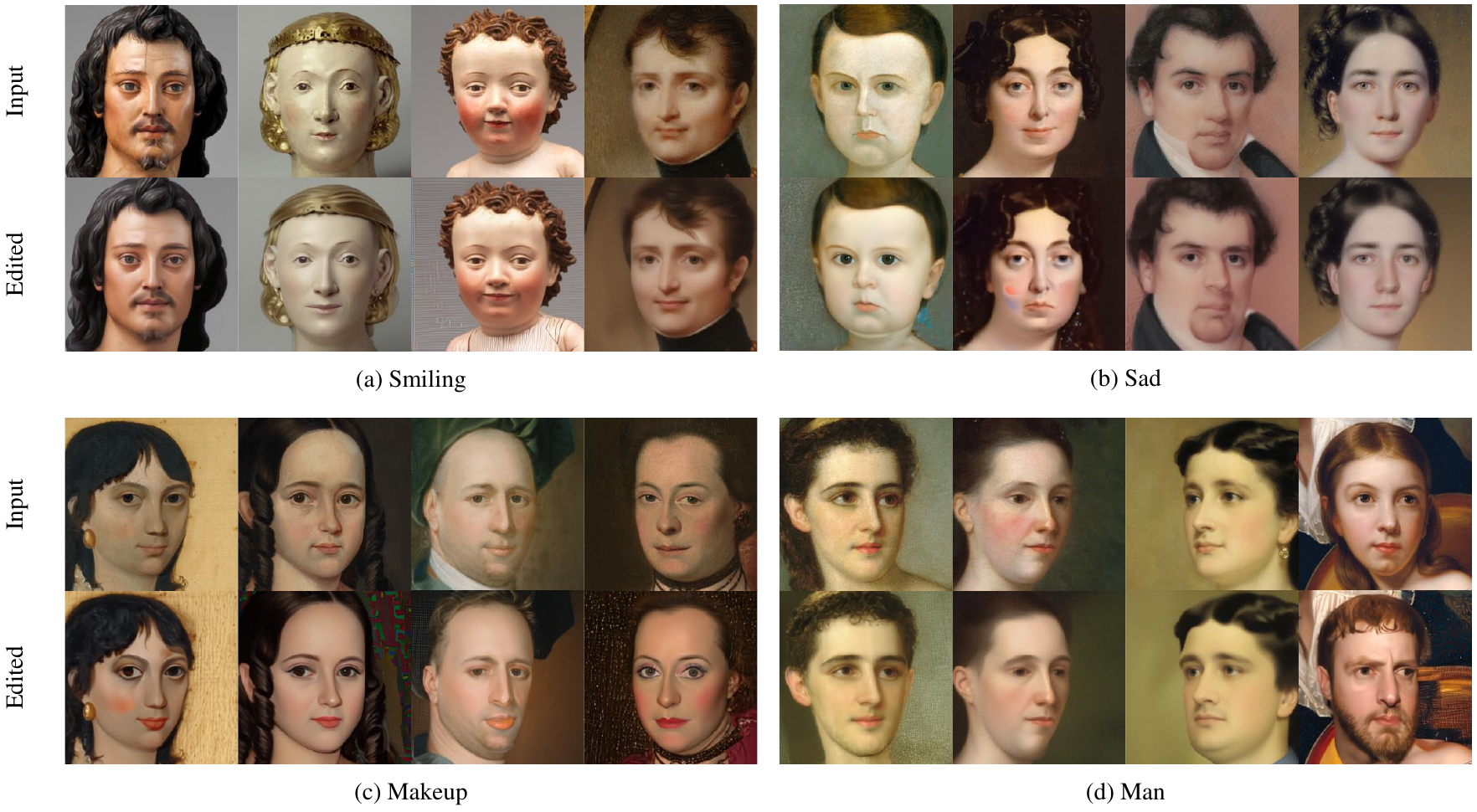}
    \caption{Applications on out-of-domain images editing on various attributes performed on METFACES \cite{karras2020training}.}
    \label{fig18}
\end{figure*}

\end{document}